\documentclass[letterpaper]{article}
\usepackage{aaai2027}
\usepackage[hyphens]{url}
\usepackage{graphicx}
\usepackage{natbib}
\usepackage{caption}
\usepackage{booktabs}
\usepackage{amsmath}
\usepackage{algorithm}
\usepackage{algpseudocode}
\usepackage{array}
\usepackage{multirow}
\usepackage{color}
\definecolor{lightgray}{RGB}{242,242,242}
\definecolor{lightblue}{RGB}{232,242,252}
\usepackage[table]{xcolor}
\definecolor{releaseblue}{RGB}{32,83,143}
\newcommand{\releasehref}[2]{\hbox{\pdfstartlink attr{/Border[0 0 0]} user{/Subtype/Link/A<</S/URI/URI(#1)>>}\textcolor{releaseblue}{#2}\pdfendlink}}
\title{Beyond Paired Inputs: Learning to Leverage RGB-T Cross-Modal Relations in Multimodal Large Language Models}

\author{Tianyi Zhao\textsuperscript{1}, Qihao Xu\textsuperscript{1}, Jiawen Xi\textsuperscript{1}, Duanni Meng\textsuperscript{1}, Bo Zhao\textsuperscript{2}, Maoxun Yuan\textsuperscript{1},\\ Xingxing Wei\textsuperscript{1}, Bo Li\textsuperscript{1}}
\affiliations{
\textsuperscript{1}School of Artificial Intelligence, Beihang University\\
\textsuperscript{2}School of Robotics and Intelligent Manufacturing, Fudan University\\
\small Code: \releasehref{https://github.com/Zhao-Tian-yi/RelMLLM}{\url{https://github.com/Zhao-Tian-yi/RelMLLM}}\quad Models: \releasehref{https://huggingface.co/tyy1850819840/RelMLLM-4B}{RelMLLM-4B}\quad\releasehref{https://huggingface.co/tyy1850819840/RelMLLM-8B}{RelMLLM-8B}
}

\newcommand{\method}{RelMLLM}
\newcommand{\methodfour}{RelMLLM-4B}
\newcommand{\methodeight}{RelMLLM-8B}

\newcommand{\IoU}{\mathrm{IoU}}

\begin{document}
\nocopyright

\maketitle

\begin{abstract}

RGB and thermal infrared (TIR) sensors provide complementary observations, yet their distinct sensing characteristics make effective cross-modal coordination challenging. Although existing multimodal large language models (MLLMs) can process paired RGB-T inputs, they often fail to translate cross-modal relations into effective task-specific decisions, leaving the potential of these relations to improve task performance and robustness largely untapped. Motivated by this observation, we introduce RelMLLM, an RGB-T MLLM that explicitly learns to exploit cross-modal relations in task-specific decision making.
It represents modality reliability, the cross-modal relation, and the dominant modality as a verifiable decision state. We derive task-grounded supervision for this state by comparing RGB-only, TIR-only, and paired RGB-T task outcomes. A hierarchical relation-guided reward supervises modality-specific evidence, relation-state prediction, and the final task output, while encouraging consistency across evidence, relation, and decision.
We evaluate RelMLLM on natural RGB-T grounding, RGB-T visual question answering, and held-out sensor degradation. RelMLLM improves grounding accuracy and degradation robustness over its Qwen3-VL baselines while preserving performance on general RGB-T reasoning benchmarks. These results demonstrate that effectively exploiting cross-modal relations leads to more accurate and robust RGB-T decisions. Code and model weights are publicly available.
\end{abstract}

\section{Introduction}

\begin{figure}[t]
\centering
\includegraphics[width=\columnwidth]{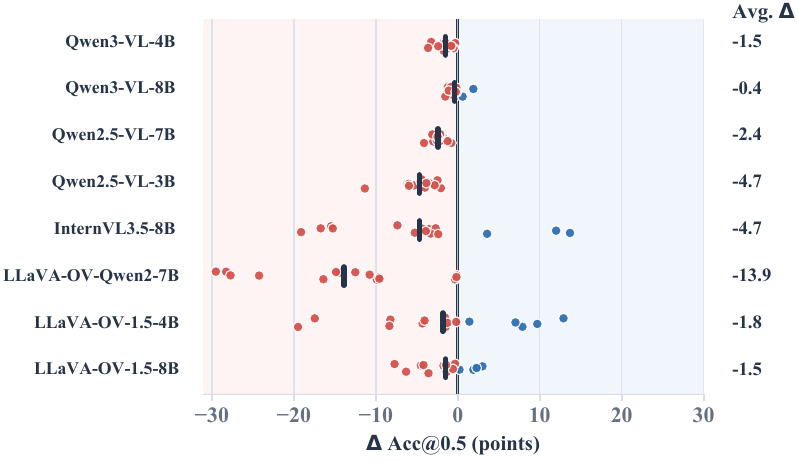}
\vspace{-6pt}
\caption{Change in RGBT-GroundBench Acc@0.5 after adding TIR to RGB for open-source MLLMs. $\Delta$ denotes RGB-T minus RGB-only. Dots represent 15 subsets.}
\vspace{-3pt}
\label{fig:intro_coordination_gap}
\end{figure}

\begin{figure*}[t]
\centering
\includegraphics[width=0.99\textwidth]{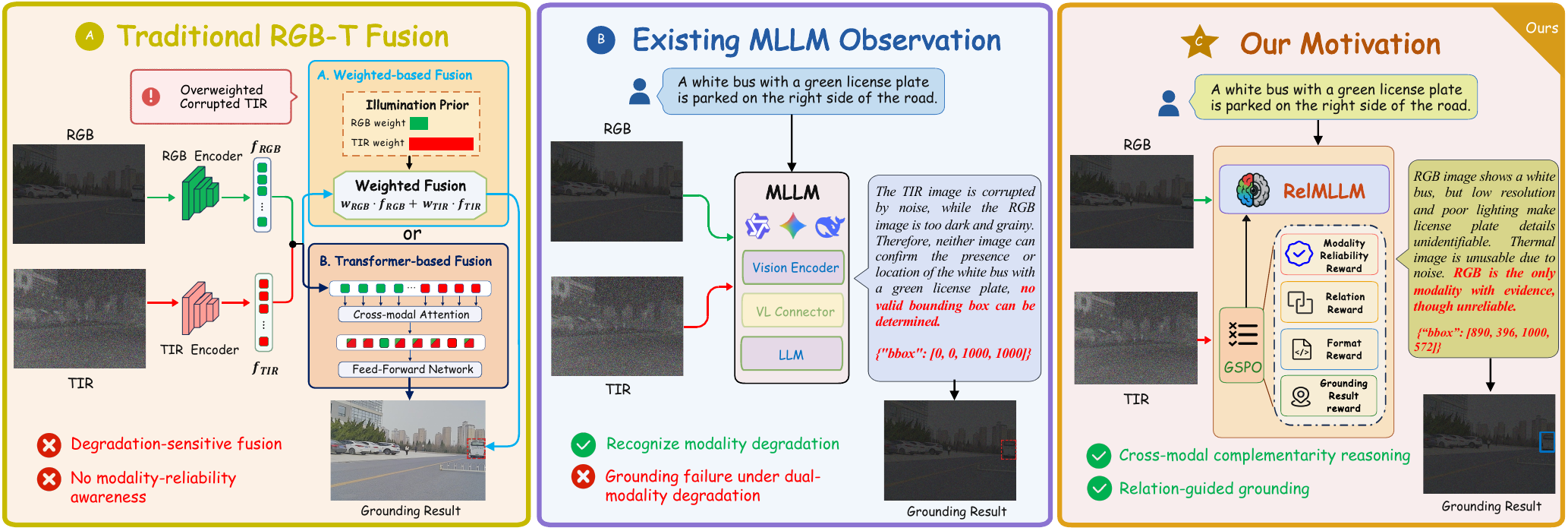}
\vspace{-3pt}
\caption{A sensor-degradation failure case. (a) Illumination-aware fusion, such as CNN-based M2D-LIF and Transformer-based IGT~\citep{zhao2025rethinking,chen2023igt}, may favor corrupted TIR. (b) A vanilla MLLM recognizes degradation but fails to respond. (c) \method{} uses cross-modal relations to localize the target from reliable evidence.}
\vspace{-3pt}
\label{fig:intro_bad_case}
\end{figure*}

RGB and thermal infrared (TIR) sensors provide complementary observations of the same scene~\citep{brenner2023rgb,gade2014thermal}. RGB images contain color, texture, and appearance details, while TIR images capture heat-related information and remain useful under poor illumination. This complementarity supports many applications, including autonomous driving~\citep{roszyk2022adopting,gonzalez2016pedestrian}, robotics~\citep{sun2019rtfnet,shivakumar2019pst900}, surveillance~\citep{gade2014thermal,lu2020cross}, and search and rescue~\citep{broyles2022wisard,gallego2019detection}.
Existing approaches to exploiting RGB-T complementarity usually combine the two modalities through specially designed CNN-~\citep{zhou2020improving,li2019illumination} or Transformer-based~\citep{shen2024icafusion,guo2024damsdet,tan2026darkact} fusion modules. Many of these methods use illumination cues to adjust the contribution of each modality, such as assigning more weight to TIR in low-light scenes. However, illumination does not always reflect sensor reliability. When illumination and sensor quality vary independently, such priors may cause unreliable evidence to dominate the final prediction.

Compared with task-specific RGB-T fusion networks~\citep{huang2026uav,ding2026adaptive}, MLLMs~\citep{majeedi2026dualvision,jiang2024mantis} provide a more general framework for using paired image observations. With large-scale vision-language pretraining and instruction-following capabilities~\citep{li2023blip,liu2023visual}, they can handle different tasks within a unified model and explicitly assess the quality and relation of multiple observations. However, the ability to process paired inputs cannot guarantee that their cross-modal relations are effectively used in task decisions. As shown in Figure~\ref{fig:intro_coordination_gap}, adding TIR reduces grounding performance in most evaluated model-subset combinations. This result shows a clear gap between accepting paired inputs and using their relations to guide the final prediction.

This gap becomes particularly clear when illumination and sensor quality provide conflicting signals. As illustrated in Figure~\ref{fig:intro_bad_case}(a), the RGB image is dark but still contains task-relevant cues, whereas the aligned TIR image is severely degraded. Illumination-driven fusion may therefore place excessive weight on the corrupted TIR observation and mislocalize the target. A vanilla MLLM can describe the poor quality of the inputs, yet it fails to recognize and use the remaining RGB evidence for localization as shown in Figure~\ref{fig:intro_bad_case}(b). This example suggests that describing sensor quality is not enough; modality reliability must be assessed with respect to the task and connected to the final decision.

Our diagnostic analysis in Section~\ref{sec:motivation} shows that relation prompting improves RGB-T grounding, but even gold relation routing remains limited when both modalities are degraded. Robust RGB-T reasoning therefore requires more than relation description or input selection: cross-modal relations must directly guide evidence use in the final prediction. This leads to our central question: \emph{how can an RGB-T MLLM learn to use sample- and task-specific cross-modal relations to guide its decisions?}

To address this challenge, we introduce \method{}, an RGB-T MLLM that organizes prediction as an evidence-relation-decision process as illustrated in Figure~\ref{fig:intro_bad_case}(c). Given paired RGB-T observations and a task instruction, it records task-relevant evidence from each observation, predicts their reliability, cross-modal relation, and dominant evidence source, and produces the task output. We derive task-grounded supervision for this process by comparing outcomes under RGB-only, TIR-only, and paired RGB-T inputs on natural data. A hierarchical relation-guided reward supervises the evidence, relation state, and task output, while encouraging consistency among them.

We evaluate \method{} on natural RGB-T grounding, RGB-T visual question answering, and a held-out sensor degradation dataset. Both model scales improve natural and degraded grounding over their Qwen3-VL initializations and achieve gains on RGB-T visual reasoning benchmarks. These results demonstrate that connecting cross-modal relations to task decisions improves both accuracy and robustness, even under sensor degradation unseen during training.
Our main contributions are summarized as follows:
\begin{itemize}
    \item We identify a key limitation of existing RGB-T MLLMs: paired inputs cannot necessarily lead to effective cross-modal coordination, and relation prompting or modality routing alone remains insufficient.

    \item We introduce \method{}, which learns a verifiable evidence-relation-decision process from automatically constructed task-grounded relation supervision and a hierarchical relation-guided reward.

    \item Extensive experiments show that \method{} improves natural RGB-T grounding and robustness to unseen controlled degradation across policy optimizers, while preserving standard RGB-T reasoning performance.
\end{itemize}

\section{Related Work}
\textbf{Task-specific RGB-thermal perception.}
RGB-T perception spans detection~\citep{hwang2015multispectral,jia2021llvip}, semantic segmentation~\citep{ha2017mfnet}, and referring grounding~\citep{zhao2025rgbt}. Existing methods align paired RGB-thermal observations and fuse or select modality features within task-specific architectures~\citep{zhang2019weakly,yuan2022translation,yuan2024c2former,zhao2025removal,yuan2025unirgbir}, while grounding models connect language to spatial predictions~\citep{kamath2021mdetr,chen2023shikra,you2024ferret,liu2024grounding,xi2026irstdagent}. However, their modality reliability and evidence-selection policies remain implicit. We instead train an MLLM to explicitly represent cross-modal relations between RGB and thermal observations and use them in downstream perceptual decisions.

\noindent\textbf{Multisensor reasoning and RGB-T MLLMs.}
Sensor-oriented benchmarks have begun to test whether MLLMs recognize and reason over thermal, depth, and other non-RGB observations. SPARK shows that explicit sensor identification can improve non-RGB reasoning~\citep{yu2024spark}, while RGB-TH-Bench evaluates RGB-thermal understanding across fine-grained abilities~\citep{moshtaghi2025rgb}. DualVision further introduces RGB-infrared instruction data and evaluates robust multimodal reasoning~\citep{majeedi2026dualvision}. These studies establish that MLLMs can process non-RGB inputs, but their primary emphasis is sensor recognition, benchmark coverage, or instruction tuning with paired RGB-thermal data. The connection between modality reliability, cross-modal relation, and the final task output remains largely implicit. Our formulation makes that connection explicit through a relation state whose components are supervised against task-grounded RGB-only, TIR-only, and paired outcomes.

\noindent\textbf{Verifiable post-training for reasoning.}
Language-model alignment has progressed from policy-gradient optimization with learned preferences to direct preference objectives and reinforcement learning with verifiable rewards~(RLVR)~\citep{schulman2017proximal,ouyang2022training,rafailov2023direct,shao2024deepseekmath,guo2025deepseek}. Group-relative optimization removes the need for a separate value model, while DAPO and GSPO refine large-scale policy updates and sequence-level optimization~\citep{yu2026dapo,zheng2025group}. Most RLVR formulations emphasize final-answer correctness and format compliance. Our method retains those verifiable signals but adds intermediate supervision for modality reliability, relation state, and evidence-to-decision consistency. It is therefore a rule-verifiable relation-guided reward design, rather than a separately trained process reward model. We evaluate GRPO, DAPO, and GSPO as optimizer variants, separating the proposed reward structure from the choice of policy optimizer.

\section{Motivation:\\ From Paired Inputs to Relation-Guided}
\label{sec:motivation}
\begin{figure}[!h]
\centering
\includegraphics[width=\columnwidth]{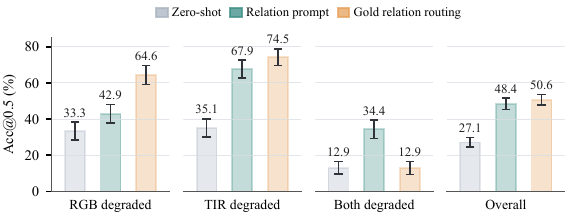}
\vspace{-3pt}
\caption{Qwen3-VL-8B on Degraded-1K. Relation prompting assesses cross-modal evidence; gold routing uses degradation labels. Error bars show 95\% bootstrap CIs.}
\vspace{-3pt}
\label{fig:degraded_legacy_diagnostic}
\end{figure}

\paragraph{Paired input does not guarantee benefit.}
Although MLLMs can process paired RGB-T inputs, they do not always benefit from the additional TIR observation. As shown in Figure~\ref{fig:intro_coordination_gap}, adding TIR reduces grounding performance in most evaluated cases, although the effect varies across models and subsets. This shows that access to paired observations does not guarantee effective use of paired evidence.

\paragraph{Relation information helps, but is not enough.}
We next examine whether relation information can improve grounding under sensor degradation.  As shown in Figure~\ref{fig:degraded_legacy_diagnostic}, a relation prompt consistently improves zero-shot grounding, showing that relation states are informative. Gold relation routing is much stronger when only one modality is degraded, but falls back to the zero-shot level when both are corrupted. Relation awareness is therefore useful, yet prompting or hard input selection alone cannot reliably exploit residual evidence distributed across two imperfect modalities.

% \paragraph{Design implication.}
Above findings motivate us to formulate RGB-T reasoning as an evidence--relation--decision process, in which cross-modal relations directly guide evidence use rather than serve only as descriptions or routing cues. Therefore, in this paper, we design \method{} to learn this process within a single paired-input generation.

\section{RelMLLM}
\label{sec:method}
\subsection{Overview and Relation-Guided Formulation}
Given an aligned RGB-T pair $(I_R,I_T)$ and a task instruction $x$, existing paired-input MLLMs typically generate the final task output $y$ directly:
\begin{equation}
p_\theta(y\mid I_R,I_T,x).
\end{equation}
In this formulation, modality reliability and cross-modal relations remain implicit. The model receives no direct guidance on when to rely on, combine, or discount each observation. \method{} instead introduces a structured relation state $z$ and models the output as
\begin{equation}
p_\theta(z,y\mid I_R,I_T,x)
=
p_\theta(z\mid I_R,I_T,x)
p_\theta(y\mid I_R,I_T,x,z).
\label{eq:relation_formulation}
\end{equation}

The main change is that $z$ must be generated before $y$ and is supervised during post-training. It describes RGB and TIR reliability, their relation, and the main evidence source. The method does not use a separate predictor or routing module. The same MLLM generates both the relation state and the final task output.

\begin{figure*}[!t]
\centering
\includegraphics[width=0.99\textwidth]{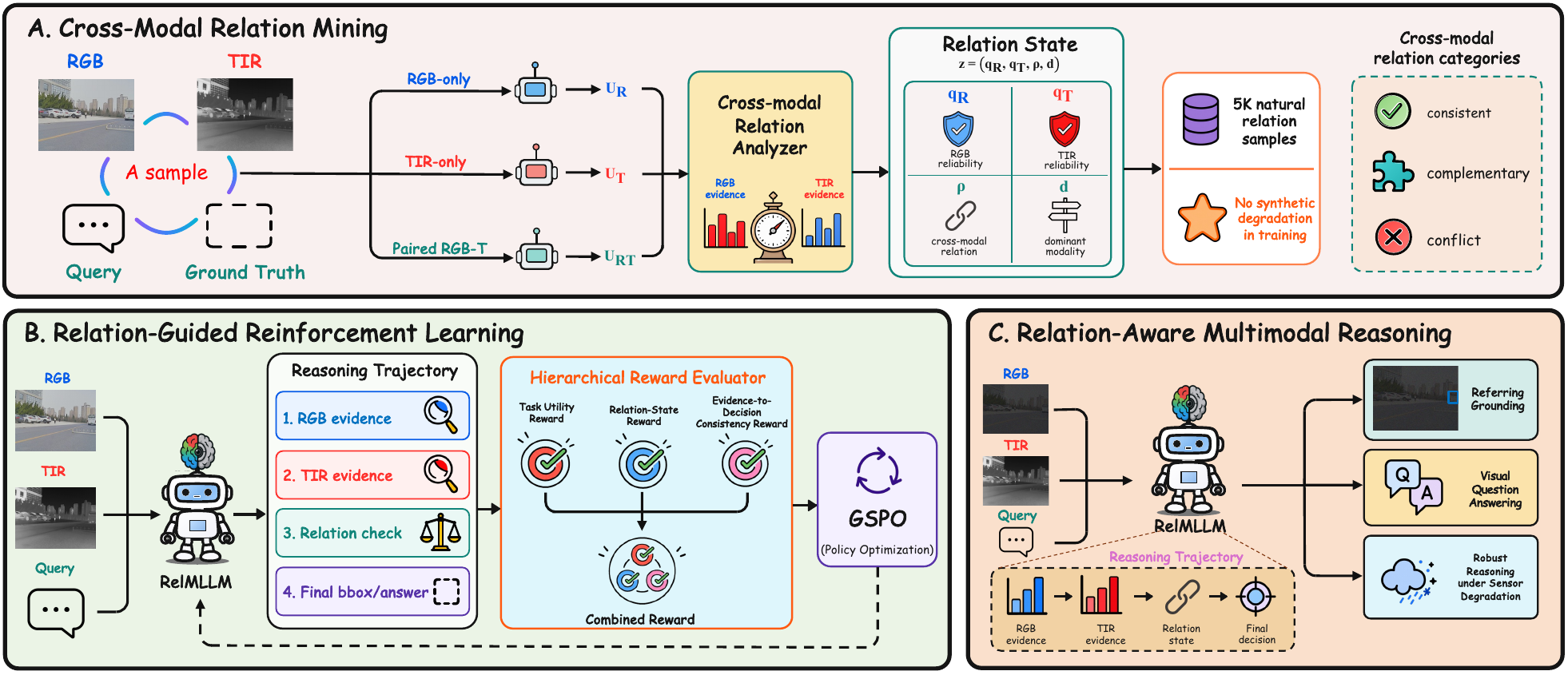}
\vspace{-2pt}
\caption{Overview of \method{}. (A) Controlled executions construct relation supervision. (B) Hierarchical rewards align evidence, relation state, and task output. (C) Inference uses one paired-input generation.}
\vspace{-2pt}
\label{fig:method_overview}
\end{figure*}

As shown in Figure~\ref{fig:method_overview}, controlled RGB-only, TIR-only, and paired executions first provide relation supervision. A hierarchical reward then evaluates evidence use, relation prediction, and task correctness. At inference, the resulting evidence-relation-decision process requires only one paired-input generation.

\begin{algorithm}[!t]
\caption{Task-Grounded Cross-Modal Relation Mining}
\label{alg:relation_mining}
\small
\begin{algorithmic}[1]
\Require Aligned pair $(I_R,I_T)$, instruction $x$, target $y^*$, frozen MLLM $f$; thresholds $\tau_h,\tau_m,\epsilon,\delta$
\Ensure Relation state $z=(q_R,q_T,\rho,d)$
\State $\hat y_R\gets f(I_R,x)$; $\hat y_T\gets f(I_T,x)$
\State $\hat y_{RT}\gets f(I_R,I_T,x)$
\For{$m\in\{R,T,RT\}$}
    \State $u_m\gets\IoU(\hat y_m,y^*)$
\EndFor
\For{$m\in\{R,T\}$}
    \State $v_m\gets\Call{ValidBox}{\hat y_m}$
    \State $q_m\gets\Call{Reliability}{u_m;\tau_h,\tau_m}$
\EndFor
\If{\Call{MisleadingPair}{$u_R,u_T,v_R,v_T;\tau_h,\epsilon$} \textbf{and} $u_{RT}<\tau_h$}
    \State $(\rho,d)\gets(\texttt{conflict},\Call{ArgMax}{u_R,u_T})$
\ElsIf{$\min(u_R,u_T)\geq\tau_h$ \textbf{and} $|u_R-u_T|\leq\delta$}
    \State $(\rho,d)\gets(\texttt{consistent},\texttt{Both})$
\ElsIf{$\max(u_R,u_T)<\tau_h$ \textbf{and} $u_{RT}\geq\tau_h$}
    \State $(\rho,d)\gets(\texttt{complementary},\texttt{Both})$
\ElsIf{$\max(u_R,u_T)\geq\tau_h$}
    \State $(\rho,d)\gets(\texttt{complementary},\Call{ArgMax}{u_R,u_T})$
\Else
    \State $(\rho,d)\gets(\texttt{uncertain},\texttt{None})$
\EndIf
\State \Return $z$
\end{algorithmic}
\end{algorithm}

\subsection{Cross-Modal Relation Mining}
\label{sec:relation_mining}

\paragraph{Relation state.}
For each task instance, we define the relation state as
\begin{equation}
z=(q_R,q_T,\rho,d),
\label{eq:relation_state}
\end{equation}
where $q_R$ and $q_T$ describe the task-specific reliability of the RGB and TIR observations as \texttt{high}, \texttt{medium}, or \texttt{low}. The relation $\rho$ is categorized as \texttt{consistent}, \texttt{complementary}, \texttt{conflict}, or \texttt{uncertain}, while $d\in\{\texttt{RGB},\texttt{TIR},\texttt{Both},\texttt{None}\}$ specifies the evidence source that should guide the decision. Together, these variables describe how the two observations support the current task rather than their intrinsic sensor quality.

\paragraph{Task-Grounded Cross-Modal Relation Mining.}
We derive relation supervision from controlled task outcomes instead of manual sensor-quality labels. For each natural RGBT-GroundBench instance, a frozen base MLLM is evaluated with RGB-only, TIR-only, and paired RGB--T inputs, producing task utilities $(u_R,u_T,u_{RT})$. For grounding, the utility is defined as the IoU between the predicted and ground-truth bboxes.

Algorithm~\ref{alg:relation_mining} converts these utilities into modality reliability, relation type, and dominant evidence source. The ordered rules distinguish consistency, complementary evidence, misleading cross-modal evidence, and cases where neither observation is reliable. In particular, \texttt{conflict} is assigned only when one modality is informative while the other produces a valid but misleading prediction that harms paired inference.
Definitions of \textsc{ValidBox}, \textsc{Reliability}, \textsc{MisleadingPair}, and thresholds are provided in the supplementary material.

The procedure yields 5K task-grounded relation annotations from natural RGB-T pairs. These annotations are outcome-derived weak labels of task-specific evidence utility rather than complete judgments of physical sensor quality. No synthetic degradation is used for annotation or training.

\subsection{Relation-Guided Reinforcement Learning}
\label{sec:relation_rl}

\paragraph{Structured reasoning trajectory.}
We represent each model response as a structured reasoning trajectory
\begin{equation}
\boldsymbol{\tau}=(e_R,e_T,e_C,\hat{z},y),
\label{eq:trajectory}
\end{equation}
where $e_R$ and $e_T$ record RGB- and TIR-specific evidence and reliability, respectively; $e_C$ records the cross-modal relation check; $\hat{z}$ is the resulting structured relation state; and $y$ is the final task output. The trajectory follows an evidence--relation--decision order: the model first examines each observation, then compares their evidence, commits to a relation state, and finally produces the task prediction. In the grounding output schema, $(e_R,e_T,e_C)$ is serialized as \texttt{steps}, an ordered, machine-verifiable decision record generated within a single response. This record enables the reward evaluator to check evidence--relation--decision consistency without treating free-form chain-of-thought as supervision. Field-level definitions are provided in the supplementary material.

\paragraph{Hierarchical reward evaluator.}
Predicting a correct relation state does not ensure that the model uses it in the final decision. We therefore design a parameter-free hierarchical reward evaluator that assigns verifiable credit at three levels: task correctness, relation-state correctness, and evidence-to-decision consistency.
The overall reward is
\begin{equation}
R=(1-\lambda_e)
\left[
(1-\lambda_r)R_{\mathrm{task}}
+\lambda_rR_{\mathrm{rel}}
\right]
+\lambda_eR_{\mathrm{cons}},
\label{eq:total_reward}
\end{equation}
where $R_{\mathrm{task}}$ evaluates the final task output, $R_{\mathrm{rel}}$ evaluates the predicted relation state, and $R_{\mathrm{cons}}$ evaluates whether the complete reasoning trajectory is consistent.
We define the consistency reward as
\begin{equation}
R_{\mathrm{cons}}
=
\alpha_sR_{\mathrm{step}}
+\alpha_mR_{\mathrm{match}}
+\alpha_dR_{\mathrm{dec}},
\label{eq:consistency_reward}
\end{equation}
where $R_{\mathrm{step}}$ checks the required reasoning order, $R_{\mathrm{match}}$ measures agreement between the predicted and reference evidence--relation records, and $R_{\mathrm{dec}}$ connects this agreement to the final task decision. Specifically, the decision-consistency term is
\begin{equation}
R_{\mathrm{dec}}
=
\IoU(y,y^*)
\left(0.5+0.5R_{\mathrm{match}}\right).
\label{eq:decision_consistency}
\end{equation}

The multiplicative design couples trajectory consistency with task correctness. If the final prediction is incorrect, a plausible relation description cannot produce a high decision-consistency reward. Conversely, a correct prediction receives full credit only when it is supported by a compatible evidence--relation trajectory. We set $\alpha_s=0.20$, $\alpha_m=0.50$, and $\alpha_d=0.30$.

\paragraph{Policy optimization.}
The hierarchical reward is independent of the policy optimizer. For each prompt, the policy samples a group of structured reasoning trajectories, which are scored by Equations~\ref{eq:total_reward}--\ref{eq:decision_consistency}. Group-relative policy updates then increase the likelihood of trajectories that jointly improve task utility, relation-state correctness, and evidence-to-decision consistency. We use GSPO for the final model and also evaluate GRPO and DAPO under the same training data, trajectory schema, and reward evaluator.

\subsection{Relation-Aware Multimodal Reasoning}

At inference, \method{} follows the evidence-relation-decision process learned during post-training. Given an aligned RGB-T pair and a task instruction, the model first identifies modality-specific evidence, then predicts the relation state, and finally produces the task output. Since the relation state appears before the final prediction in the autoregressive response, it provides an explicit intermediate representation for deciding whether to rely on, combine, or discount evidence from the two observations. All components are generated by the same MLLM within one structured response. The RGB-only and TIR-only executions used for relation mining, as well as the hierarchical reward evaluator, are required only during training and introduce no additional model calls at inference.

\section{Experiments}
\subsection{Experimental Setup}

\paragraph{Datasets.}
Post-training uses natural RGB-T pairs from RGBT-GroundBench, with a 5,000-example outcome-labeled subset for relation-guided reinforcement learning (RL). Neither SFT nor RL uses synthetic degradation. We evaluate grounding on RGBT-GroundBench~\citep{zhao2025rgbt} and Degraded-1K, and visual question answering (VQA) on RGB-TH-Bench~\citep{moshtaghi2025rgb} and DV-1K-Balanced. DV-1K-Balanced combines the original 500 RGB-category DV-500 examples~\citep{majeedi2026dualvision} with 500 held-out infrared-category examples, while Degraded-1K contains unseen RGB, TIR, and joint corruptions. Table~\ref{tab:evaluation_sets} summarizes the evaluation splits.
\begin{table}[!t]
\centering
\footnotesize
\setlength{\tabcolsep}{2.6pt}
\renewcommand{\arraystretch}{0.9}
\begin{tabular}{lccc}
\toprule
Benchmark & Test Samples $N$ & Task & Metric \\
\midrule
RGBT-GroundBench & 10,124 & Ground. & Acc@0.5 \\
RGB-TH-Bench & 1,856 & VQA & QAcc \\
DV-1K. & 1,000 & VQA & Acc.  \\
Degraded-1K & 1,000 & Ground. & Acc@0.5 \\
\bottomrule
\end{tabular}
\normalsize
\vspace{-3pt}
\caption{Evaluation benchmarks and reporting protocols.}
\vspace{-3pt}
\label{tab:evaluation_sets}
\end{table}
\paragraph{Evaluation and implementation details.}
Grounding uses original-resolution Acc@0.5; GroundBench Overall covers 10,124 mutually exclusive all-test queries and excludes the overlapping testA/B/C diagnostics. RGB-TH-Bench reports macro-averaged RGB-text and RGB-thermal-text QAcc and their mean, while DV-1K-Balanced uses answer-type-aware accuracy over its RGB and infrared subsets; invalid outputs are incorrect. Our 4B/8B models initialize from Qwen3-VL-Instruct and use LoRA ($r=8$, $\alpha=16$, dropout 0.05). RL runs for one epoch at $1{\times}10^{-6}$ with four samples per prompt and a 1,024-token limit. Training uses bfloat16 on two A100 GPUs; evaluation uses deterministic decoding.

\subsection{Ablation Studies}
\paragraph{Relation guidance across policy optimizers.}
Table~\ref{tab:optimizer_ablation} disentangles the effect of relation guidance from the choice of policy optimizer. Across GRPO, DAPO, and GSPO, relation-guided training consistently improves Degraded-1K Overall by about 30 points, while the three variants differ by only 0.60 points, indicating that the robustness gain primarily stems from the proposed objective rather than a specific RL method. Performance on standard RGB-T benchmarks remains largely preserved. We use GSPO for the final model because it achieves the strongest degradation robustness.

\begin{table*}[t]
\centering
\footnotesize
\setlength{\tabcolsep}{5.2pt}
\renewcommand{\arraystretch}{0.9}
\begin{tabular}{lrrrrrr}
\toprule
\multirow[c]{2}{*}[-0.8ex]{\textbf{Method}} &
\multirow[c]{2}{*}[-0.8ex]{\shortstack[c]{\textbf{RGB-TH-Bench}\\\textbf{Overall QAcc (\%)}}} &
\multirow[c]{2}{*}[-0.8ex]{\shortstack[c]{\textbf{DV-1K}\\\textbf{Overall Acc. (\%)}}} &
\multicolumn{4}{c}{\textbf{Degraded-1K Acc@0.5 (\%)}} \\
\cmidrule(lr){4-7}
& & & \textbf{RGB deg.} & \textbf{TIR deg.} & \textbf{Both deg.} & \textbf{Overall} \\
\midrule

Vanilla GRPO-only & 74.41 & 78.10 & 31.23 & 48.65 & 12.57 & 30.80 \\
\rowcolor{lightgray}Relation-GRPO~(Ours)
& 75.50~({+1.09})
& 78.10~({+0.00})
& 57.06~({+25.83})
& 74.17~({+25.52})
& 50.30~({+37.73})
& 60.50~({+29.70}) \\
\midrule
Vanilla DAPO-only & 74.52 & 78.10 & 30.03 & 48.95 & 13.17 & 30.70 \\
\rowcolor{lightgray}Relation-DAPO~(Ours)
& 75.38~({+0.86})
& \underline{\textbf{78.60}}~({+0.50})
& \underline{\textbf{57.66}}~({+27.63})
& 74.17~({+25.22})
& 50.60~({+37.43})
& 60.80~({+30.10}) \\
\midrule
Vanilla GSPO-only & 75.49 & 77.80 & 32.73 & 48.05 & 14.37 & 31.70 \\
\rowcolor{lightgray}Relation-GSPO~(Ours)
& \underline{\textbf{75.61}}~({+0.12})
& 78.30~({+0.50})
& 57.36~({+24.63})
& \underline{\textbf{74.47}}~({+26.42})
& \underline{\textbf{51.50}}~({+37.13})
& \underline{\textbf{61.10}}~({+29.40}) \\
\bottomrule
\end{tabular}
\normalsize
\vspace{-3pt}
\caption{Task-only versus relation-guided RL with Qwen3-VL-8B. Scores are percentages; parentheses show absolute gains over the corresponding task-only optimizer in percentage points.}
\vspace{-3pt}
\label{tab:optimizer_ablation}
\end{table*}

\begin{table*}[!t]
\centering
\footnotesize
\setlength{\tabcolsep}{11.5pt}
\renewcommand{\arraystretch}{0.9}
\begin{tabular}{lcccccc}
\toprule
\multirow[c]{2}{*}[-0.8ex]{\textbf{Variant}} &
\multirow[c]{2}{*}[-0.8ex]{\shortstack{\textbf{RGB-TH-Bench}\\\textbf{Overall QAcc (\%)}}} &
\multirow[c]{2}{*}[-0.8ex]{\shortstack{\textbf{DV-1K}\\\textbf{Overall Acc. (\%)}}} &
\multicolumn{4}{c}{\textbf{Degraded-1K Acc@0.5 (\%)}} \\
\cmidrule(lr){4-7}
& & & \textbf{RGB deg.} & \textbf{TIR deg.} & \textbf{Both deg.} & \textbf{Overall}\\
\midrule
Zero-shot & 73.40 & 77.4 & 33.33 & 35.14 & 12.87 & 27.10 \\
SFT only & 73.04 & 77.7 & 43.84 & 52.85 & 38.62 & 45.10 \\
w/o Relation-State & 75.44 & 78.2 & 45.05 & 55.86 & 41.02 & 47.30 \\
w/o Evidence Cons. & 75.50 & \underline{\textbf{78.6}} & 55.56 & 72.07 & 47.90 & 58.50 \\
\midrule
\rowcolor{lightgray}\textbf{\methodeight{}} & \underline{\textbf{75.61}} & 78.3 & \underline{\textbf{57.36}} & \underline{\textbf{74.47}} & \underline{\textbf{51.50}} & \underline{\textbf{61.10}} \\
\bottomrule
\end{tabular}
\normalsize
\vspace{-3pt}
\caption{Ablation of relation-guided post-training components on Qwen3-VL-8B. ``w/o Relation-State'' and ``w/o Evidence Cons.'' remove $R_{\mathrm{rel}}$ and $R_{\mathrm{cons}}$ of Eq.~\ref{eq:total_reward}, respectively; all other settings are held fixed.}
\vspace{-3pt}
\label{tab:main_reward_ablation}
\end{table*}

\begin{table}[!t]
\centering
\footnotesize
\setlength{\tabcolsep}{4.6pt}
\renewcommand{\arraystretch}{0.9}
\begin{tabular}{lccc}
\toprule
\multirow[c]{2}{*}[-0.8ex]{\textbf{Model}} & \multicolumn{3}{c}{\textbf{RGB-TH-Bench  QAcc. (\%)}} \\ \cmidrule{2-4}
& \shortstack{\textbf{RGB-Txt}} & \shortstack{\textbf{RGB-Th-Txt}} & \textbf{Overall} \\
\midrule
Random Baseline & 50.00 & 50.00 &  50.00 \\
\midrule
InternVL3-2B & 77.59 & 59.74 & 68.67 \\
Qwen2.5-VL-3B & 78.93 & 57.62 & 68.28 \\
Qwen3-VL-4B & 82.95 & 63.13 & 73.04 \\
InternVL2.5-4B & 78.26 & 59.15 & 68.70 \\
LLaVA-OV-1.5-4B& 79.89 & 57.28 & 68.58 \\
LLaVA-OV-Qwen2-7B & 80.75 & 62.58 & 71.66 \\
Qwen2.5-VL-7B & 78.93 & 66.76 & 72.85 \\
InternVL3-8B & 82.28 & 55.55 & 68.92 \\
LLaVA-OV-1.5-8B & 82.66 & 55.67 & 69.17 \\
InternVL2.5-8B\textsuperscript{\dag} & 79.68 & 63.08 & 71.38 \\
MiniCPM-V-2.6\textsuperscript{\dag} & 79.80 & 63.31 & 71.56 \\
MiniCPM-o-2.6\textsuperscript{\dag} & 81.16 & 62.20 & 71.68 \\
Qwen3-VL-8B & 83.72 & 63.02 & 73.40 \\
GPT-4o\textsuperscript{\dag} & 84.11 & 60.85 & 72.48 \\
\midrule
\rowcolor{lightgray}\methodfour{} & 82.18 & \underline{\textbf{68.74}} & 75.46 \\
\rowcolor{lightgray}\methodeight{} & \underline{\textbf{84.58}} & 66.64 & \underline{\textbf{75.61}} \\
\bottomrule
\end{tabular}
\normalsize
\vspace{-3pt}
\caption{RGB-TH-Bench QAcc (\%). \textsuperscript{\dag}quoted from~\cite{moshtaghi2025rgb}.}
\vspace{-3pt}
\label{tab:rgbth_summary}
\end{table}

\paragraph{Contribution of post-training components.}
Table~\ref{tab:main_reward_ablation} isolates the effects of SFT, relation-state supervision, and evidence-to-decision consistency. The substantial gap between SFT and the full model shows that task adaptation alone is insufficient for robust cross-modal evidence utilization. Removing $R_{\mathrm{rel}}$ causes the largest degradation, demonstrating that explicit relation-state supervision is the primary driver of robustness, while the additional drop without $R_{\mathrm{cons}}$ confirms the benefit of coupling relation assessment with the final decision. Standard RGB-T benchmark performance remains stable, indicating that these robustness gains do not come at the cost of general reasoning ability.

\subsection{Results on Natural RGB-T Benchmarks}

\paragraph{RGB-T Visual Question Answering.}
Tables~\ref{tab:rgbth_summary} and~\ref{tab:dv1k_summary} examine whether relation-guided post-training transfers beyond grounding while preserving general RGB-T reasoning ability. On RGB-TH-Bench, both model scales improve Overall QAcc, with the gains primarily concentrated in RGB-thermal-text reasoning; in particular, \methodfour{} improves this group from 63.13 to 68.74. This suggests that explicit relation supervision strengthens the use of thermal evidence rather than merely adapting the model to grounding outputs. On DV-1K-Balanced, performance is largely preserved or modestly improved, with \methodfour{} achieving the best overall result and \methodeight{} improving its initialization from 77.4 to 78.3, a gain of 0.9 percentage points. Taken together, these results show that the substantial robustness gains of \method{} do not come at the cost of standard RGB-T reasoning performance.

\begin{table}[!t]
\centering
\footnotesize  
\setlength{\tabcolsep}{3.8pt}
\renewcommand{\arraystretch}{0.9}
\begin{tabular}{lccc}
\toprule
\multirow[c]{2}{*}[-0.8ex]{\textbf{Model}} & \multicolumn{3}{c}{\textbf{DV-1K  Acc. (\%)}} \\ \cmidrule{2-4}

& \shortstack{\textbf{DV-RGB~(DV-500)}} & \shortstack{\textbf{DV-TIR}} & \shortstack{\textbf{Overall}} \\
% \midrule
% LLaVA-1.5 & 81.20 & -- & -- \\
% LLaMA-4-Scout & 85.80 & -- & -- \\
% Claude-Opus-4 & 86.60 & -- & -- \\
% Claude-Sonnet-3.5-v2 & 87.40 & -- & -- \\
% Qwen2-VL & 89.80 & -- & -- \\
% Qwen2.5-VL & 90.20 & -- & -- \\
\midrule
InternVL2.5-4B & 87.2 & 58.4 & 72.8 \\
InternVL3-2B & 88.4 & 56.0 & 72.2 \\
InternVL3-8B & 88.0 & 67.8 & 77.9 \\
LLaVA-OV-1.5-4B & 88.2 & 66.6 & 77.4 \\
LLaVA-OV-1.5-8B & 86.0 & \underline{\textbf{68.0}} & 77.0 \\

Qwen3-VL-4B & 90.6 & 66.4 & 78.5 \\
Qwen3-VL-8B & 87.2 & 67.6 & 77.4 \\
DualVision\textsuperscript{\dag} & 88.4 & -- & -- \\
\midrule
\rowcolor{lightgray}\methodfour{} & \underline{\textbf{92.6}} & 66.0 & \underline{\textbf{79.3}} \\
\rowcolor{lightgray}\methodeight{} & 89.2 & 67.4 & 78.3 \\
\bottomrule
\end{tabular}
\normalsize
\vspace{-3pt}
\caption{DV-1K-Balanced accuracy (\%). DV-RGB is the original DV-500 split; \textsuperscript{\dag}quoted from~\cite{majeedi2026dualvision}.}
\vspace{-3pt}
\label{tab:dv1k_summary}
\end{table}

\begin{table}[!t]
\centering
\footnotesize
\setlength{\tabcolsep}{2.5pt}
\renewcommand{\arraystretch}{0.9}
\begin{tabular}{lcccc}
\toprule
\multirow[c]{2}{*}[-0.8ex]{\textbf{Model}} & \multicolumn{4}{c}{\textbf{Degraded-1K Acc@0.5 (\%)}} \\ \cmidrule{2-5}
 & \textbf{RGB deg.} & \textbf{TIR deg.} & \textbf{Both deg.} & \textbf{Overall} \\
\midrule
Qwen3-VL-4B& 8.71& 18.02& 22.46& 16.40 \\
\rowcolor{lightgray}
& \textbf{{53.15}}
& \textbf{{64.86}}
& \textbf{{28.14}}
& \textbf{{48.70}} \\
\rowcolor{lightgray} \multirow{-2}{*}{\methodfour{}}&{(+44.44)}&{(+46.84)}&{(+5.68)}&{(+32.30)} \\
\midrule
Qwen3-VL-8B& 33.33& 35.14& 12.87& 27.10 \\
\rowcolor{lightgray}  & \textbf{57.36} & {\textbf{74.47}} & \textbf{51.50} & {\textbf{61.10}} \\
\rowcolor{lightgray} \multirow{-2}{*}{\methodeight{}} &{(+24.03)} &{(+39.33)}&{(+38.63)}&{(+34.00)} \\
\bottomrule
\end{tabular}%
\normalsize
\vspace{-3pt}
\caption{Acc@0.5 (\%) on Degraded-1K.}
\vspace{-3pt}
\label{tab:degraded_breakdown}
\end{table}

\definecolor{sEightOpen}{RGB}{231,224,249}
\definecolor{sEightClosed}{RGB}{220,233,250}
\definecolor{sEightRGBT}{RGB}{222,239,234}
\definecolor{sEightVLM}{RGB}{244,236,218}
\definecolor{sEightOurs}{RGB}{247,243,252}
\newcommand{\sEightIcon}[1]{%
  \raisebox{-0.34em}{\includegraphics[height=1.45em]{#1}}\hspace{2pt}}

\begin{table*}[!t]
\centering
\scriptsize
\setlength{\tabcolsep}{2pt}
\renewcommand{\arraystretch}{0.9}
\resizebox{\textwidth}{!}{%
\begin{tabular}{lcc|ccccc|ccccc|ccccc}
\toprule
\multirow[c]{2}{*}{\textbf{Model}} &
\multirow[c]{2}{*}{\textbf{Rank}} &
\multirow[c]{2}{*}[-0.8ex]{\shortstack[c]{\textbf{Test}\\\textbf{Overall}}} &
% \multirow{2}{*}{Model} & \multirow{2}{*}{Rank} & \multirow{2}{*}{\shortstack{Test\\Overall}} &
\multicolumn{5}{c|}{\textbf{RefFLIR Acc@0.5 (\%)}} &
\multicolumn{5}{c|}{\textbf{RefM3FD Acc@0.5 (\%)}} &
\multicolumn{5}{c}{\textbf{RefMFAD Acc@0.5 (\%)}} \\
\cmidrule(lr){4-8}\cmidrule(lr){9-13}\cmidrule(lr){14-18}
& & & \textbf{val} & \textbf{test}& \textbf{testA} & \textbf{testB} & \textbf{testC} & \textbf{val} & \textbf{test}& \textbf{testA} & \textbf{testB} & \textbf{testC}& \textbf{val} & \textbf{test}& \textbf{testA} & \textbf{testB} & \textbf{testC} \\
\midrule
% \rowcolor{sEightVLM}
\rowcolor{gray!15}
\multicolumn{18}{c}{\textbf{\textit{(A) VLM-Based RGB-T Visual Grounding Models}}} \\
MV-CLIP-VG\textsuperscript{\dag} & 16 & 44.00 & 45.57 & 46.01 & 72.62 & 41.56 & 16.06 & 34.52 & 38.92 & 63.42 & 50.05 & 10.01 & 47.02 & 47.52 & 82.76 & 45.07 & 23.96 \\
MV-HiVG-B\textsuperscript{\dag} & 8 & 69.47 & \underline{\textbf{75.33}} & 69.20 & 85.80 & 66.72 & 50.20 & 69.64 & 72.35 & 93.99 & 79.43 & 49.40 & 67.07 & 67.04 & 91.01 & 64.52 & 51.53 \\
MV-OneRef-B\textsuperscript{\dag} & 9 & 61.90 & 63.82 & 62.05 & 86.19 & 57.34 & 35.43 & 62.50 & 63.98 & 89.20 & 69.34 & 37.93 & 61.62 & 59.97 & 89.25 & 56.85 & 41.58 \\
UniRGB-IR & 7 & 69.89 & 70.23 & 69.87 & 90.32 & 59.06 & 47.46 & 71.43 & 73.07 & 94.40 & 77.06 & 51.89 & 68.07 & 67.06 & 90.49 & 64.11 & 50.71 \\
RGBT-VGNet\textsuperscript{\dag} & 3 & 71.31 & \underline{\textbf{75.33}} & 72.43 & 89.64 & 69.22 & 52.53 & \underline{\textbf{73.21}} & 74.89 & 94.16 & \underline{\textbf{80.75}} & 55.52 & \underline{\textbf{68.31}} & 67.58 & \underline{\textbf{91.29}} & 64.80 & 51.18 \\
\midrule
% \rowcolor{sEightOpen}
\rowcolor{gray!15}
\multicolumn{18}{c}{\textbf{\textit{(B) Open-Source Multimodal Large Language Models}}} \\

\sEightIcon{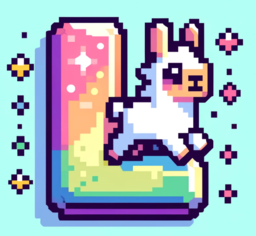}LLaVA-OV-1.5-8B & 19 & 1.39 & 3.45 & 4.47 & 7.68 & 4.43 & 0.31 & 1.79 & 0.82 & 1.22 & 1.01 & 0.27 & 0.08 & 0.38 & 0.38 & 0.33 & 0.31 \\
\sEightIcon{figures/model_icons/llava.png}LLaVA-OV-Qwen2-7B & 18 & 1.75 & 3.29 & 3.37 & 6.69 & 2.34 & 0.00 & 1.19 & 2.57 & 4.46 & 3.56 & 0.16 & 1.51 & 0.21 & 0.38 & 0.20 & 0.04 \\
\sEightIcon{figures/model_icons/llava.png}LLaVA-OV-1.5-4B & 17 & 6.16 & 14.80 & 15.92 & 28.55 & 13.13 & 1.52 & 4.76 & 7.18 & 6.41 & 16.54 & 0.87 & 0.24 & 0.42 & 0.89 & 0.37 & 0.20 \\
\sEightIcon{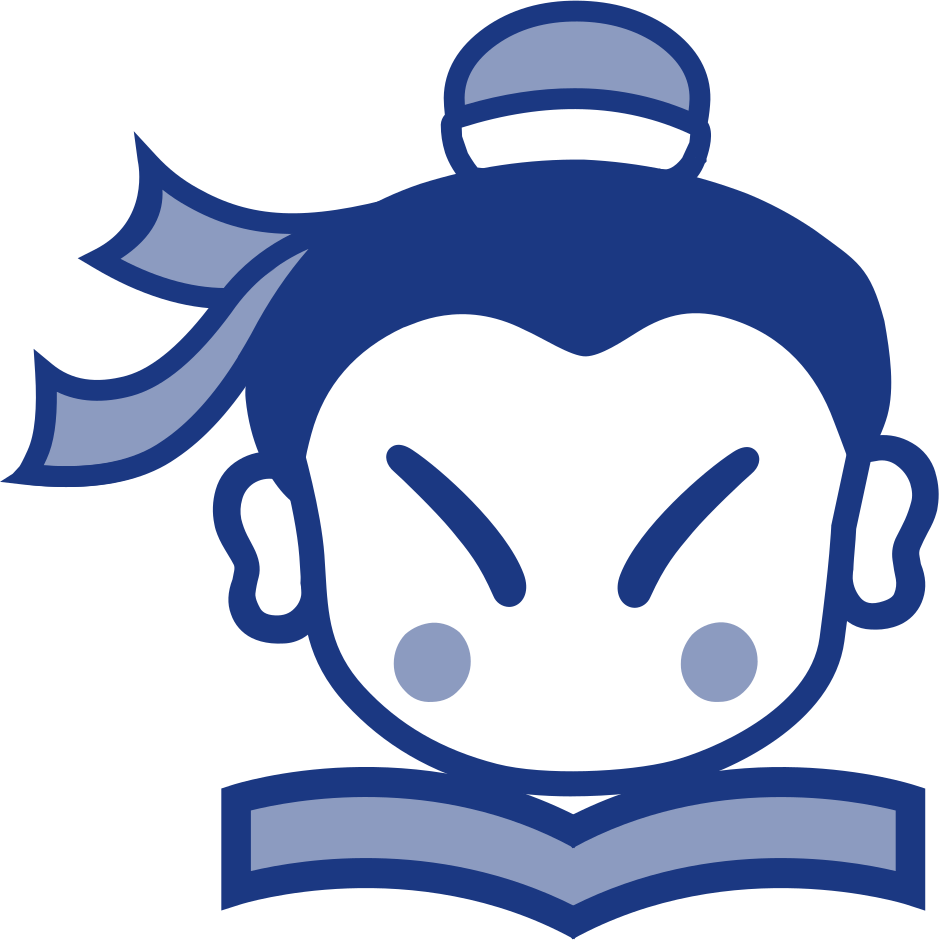}InternVL3.5-8B & 14 & 53.46 & 40.62 & 46.53 & 62.79 & 51.74 & 23.45 & 62.50 & 66.37 & 88.64 & 68.83 & 44.70 & 41.24 & 45.41 & 65.57 & 43.38 & 31.38 \\
\sEightIcon{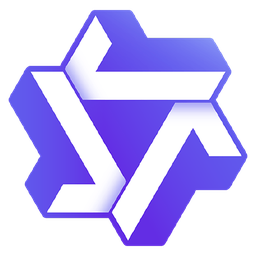}Qwen2.5-VL-3B & 13 & 53.60 & 57.07 & 64.64 & 86.55 & 50.79 & 42.15 & 54.17 & 53.28 & 81.66 & 51.01 & 28.30 & 49.20 & 48.40 & 69.11 & 46.24 & 33.58 \\
\sEightIcon{figures/model_icons/qwen.png}Qwen2.5-VL-7B & 10 & 60.66 & 60.69 & 72.53 & 91.36 & 61.08 & 53.51 & 59.52 & 59.83 & 88.56 & 63.25 & 32.58 & 55.10 & 55.51 & 80.00 & 52.17 & 39.08 \\
\sEightIcon{figures/model_icons/qwen.png}Qwen3-VL-4B & 6 & 70.31 & 63.49 & 72.43 & 90.40 & 66.61 & 52.58 & 69.64 & 73.65 & 92.78 & 73.49 & 55.63 & 66.32 & 66.28 & 85.55 & 63.45 & 53.14 \\
\sEightIcon{figures/model_icons/qwen.png}Qwen3-VL-8B & 5 & 70.84 & 62.99 & 74.76 & 92.20 & 69.94 & 54.86 & 72.02 & 74.23 & 94.40 & 74.59 & 54.38 & 65.84 & 65.88 & 86.19 & 63.16 & 51.85 \\

\midrule
\rowcolor{gray!15}
% \rowcolor{sEightClosed}
\multicolumn{18}{c}{\textbf{\textit{(C) Proprietary Multimodal Large Language Models}}} \\
\sEightIcon{figures/model_icons/qwen.png}Qwen3-VL-Plus & 15 & 49.67 & 63.82 & 73.15 & 91.48 & 71.84 & 51.76 & 42.26 & 53.58 & 69.07 & 71.30 & 30.68 & 35.27 & 34.54 & 61.14 & 33.03 & 17.67 \\
\sEightIcon{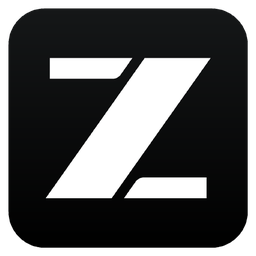}GLM-4.6V-Flash & 12 & 54.92 & 53.12 & 56.84 & 76.59 & 49.53 & 35.74 & 56.55 & 59.00 & 80.28 & 61.43 & 39.34 & 49.12 & 50.33 & 73.04 & 48.65 & 33.78 \\
\sEightIcon{figures/model_icons/qwen.png}Qwen3.5-Plus & 11 & 59.45 & 62.66 & 71.53 & 92.08 & 63.29 & 48.76 & 53.57 & 60.25 & 77.84 & 71.57 & 37.61 & 53.50 & 52.76 & 73.04 & 50.90 & 38.53 \\
\sEightIcon{figures/model_icons/glm.png}GLM-4.6V & 4 & 71.18 & 67.43 & 76.76 & \underline{\textbf{93.64}} & 70.25 & \underline{\textbf{58.68}} & 72.02 & 75.03 & 93.02 & 77.06 & 57.36 & 65.29 & 64.99 & 84.30 & 62.22 & 51.73 \\
\midrule
\rowcolor{gray!15}
\multicolumn{18}{c}{\textbf{\textit{(D) RGB-T Multimodal Large Language Models}}} \\
% \rowcolor{lightgray}
\sEightIcon{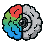}\textbf{\methodfour{}} & 2 & 72.36 & 62.83 & 73.95 & 93.28 & 67.72 & 52.79 & 67.86 & 76.11 & 93.75 & 74.95 & \underline{\textbf{59.15}} & 65.84 & 68.24 & 86.31 & 65.49 & 55.66 \\
% \rowcolor{lightgray}
\sEightIcon{figures/model_icons/unirgbir.png}\textbf{\methodeight{}} & \textbf{1} & \underline{\textbf{73.30}} & 66.94 & \underline{\textbf{77.00}} & {93.16} & \underline{\textbf{72.47}} & {58.06} & 71.43 & \underline{\textbf{76.22}} & \underline{\textbf{94.97}} & {77.42} & {57.95} & 67.68 & \underline{\textbf{68.87}} & 86.58 & \underline{\textbf{66.48}} & \underline{\textbf{56.36}} \\
\bottomrule
\end{tabular}
}
\normalsize
\vspace{-2pt}
\caption{Results on RGBT-GroundBench. Test Overall is across the three full test splits and excludes val and testA/B/C. Rank is determined globally by Test Overall. \textsuperscript{\dag}quoted from~\cite{zhao2025rgbt}.}
\vspace{-2pt}
\label{tab:paired_mllm_15subset_acc50}
\end{table*}

\paragraph{RGB-T Visual Grounding.}
To evaluate natural paired-input grounding, Table~\ref{tab:paired_mllm_15subset_acc50} compares \method{} with task-specific RGB-T grounding models, open-source general MLLMs, and proprietary MLLMs across the three sources and 15 diagnostic subsets of RGBT-GroundBench. Both model scales consistently improve over their Qwen3-VL initializations, showing that relation-guided post-training benefits natural RGB-T grounding beyond degradation robustness. More importantly, \methodeight{} achieves the best Overall Acc@0.5 of 73.30, surpassing the strongest task-specific model, RGBT-VGNet, by 1.99 points and the strongest proprietary MLLM, GLM-4.6V, by 2.12 points. These results demonstrate that explicitly learning to exploit cross-modal relations enables a general MLLM to outperform both specialized RGB-T architectures and stronger general-purpose MLLMs on paired-input grounding. More results are provided in the supplementary material.
% \begin{table}[t]
% \centering
% \footnotesize
% \setlength{\tabcolsep}{0.8pt}
% \begin{tabular}{@{}llrrrr@{}}
% \toprule
% Model & Type & FLIR & M3FD & MFAD & Overall \\
% \midrule
% Qwen3-VL-Plus & API & 73.15 & 53.58 & 34.54 & 49.67 \\
% Qwen2.5-VL-3B & OSS & 64.64 & 53.28 & 48.40 & 53.60 \\
% GLM-4.6V-Flash & API & 56.84 & 59.00 & 50.33 & 54.92 \\
% Qwen3.5-Plus & API & 71.53 & 60.25 & 52.76 & 59.45 \\
% Qwen2.5-VL-7B & OSS & 72.53 & 59.83 & 55.51 & 60.66 \\
% Qwen3-VL-8B & OSS & 74.76 & 74.23 & 65.88 & 70.84 \\
% GLM-4.6V & API & 76.76 & 75.03 & 64.99 & 71.18 \\
% Qwen3-VL-4B & OSS & 72.43 & 73.65 & 66.28 & 70.31 \\
% \methodfour{} & Ours & 73.95 & 76.11 & 68.24 & 72.36 \\
% \textbf{\methodeight{}} & Ours & \textbf{77.00} & \textbf{76.22} & \textbf{68.87} & \textbf{73.30} \\
% \bottomrule
% \end{tabular}
% \normalsize
% \caption{RGB-T grounding Acc@0.5 (\%) on RGBT-GroundBench. Overall is sample-count weighted; API denotes closed-source inference.}
% \label{tab:groundbench_main}
% \end{table}

\subsection{Robustness under Controlled Sensor Degradation}

Table~\ref{tab:degraded_breakdown} evaluates robustness to held-out RGB, TIR, and joint corruptions. Degraded-1K is used exclusively for evaluation, and neither its images nor corruption configurations are observed during post-training. Both model scales substantially outperform their Qwen3-VL initializations, with \methodfour{} gaining 32.30 points Overall and \methodeight{} improving from 27.10 to 61.10. The most pronounced gain occurs under joint degradation, where the 8B model improves from 12.87 to 51.50. This setting particularly probes the intended capability of \method{}: exploiting residual evidence distributed across two imperfect observations, rather than simply selecting a nominally preferred modality.

The 4B model nevertheless exhibits a scale-dependent failure mode under joint degradation. Although \methodfour{} improves over its initialization from 22.46 to 28.14, it remains substantially behind \methodeight{}. An output audit shows that 201 of 334 responses fail to produce a valid bounding box and are frequently associated with low predicted reliability, an \texttt{uncertain} relation, or a \texttt{None} dominant-modality prediction. An invalid bounding box alone does not establish an explicit refusal. This suggests that, for the 4B model, uncertainty estimation is not always converted into a valid best-effort prediction. The result points to a remaining challenge of better coupling relation-state prediction with final-task execution in compact RGB-T MLLMs.

% \section{Discussion}
% Our results separate modality identity, relation awareness, and relation utilization. Prompt-matched controls show that elicitation helps, while post-training adds gains under direct and structured interfaces. Transfer to held-out corruption is notable because training uses only natural pairs. However, outcome-derived labels can inherit base-model errors, conflict is rare, and synthetic corruption cannot represent every hardware failure. Near-constant RGB-TH-Bench results bound the claim to relation-conditioned grounding. Additional limitations and failure cases appear in the supplementary material.

\section{Conclusion}
We presented \method{}, an RGB-T MLLM that explicitly connects cross-modal relation assessment with task-specific decision making. Through automatically constructed task-grounded relation supervision and a hierarchical relation-guided reward, the model learns to assess modality reliability, reason about cross-modal relations, and use the resulting evidence in its final prediction. Experiments on natural RGB-T grounding, visual question answering, and held-out controlled degradation show that \method{} improves grounding accuracy and substantially enhances degradation robustness while preserving general RGB-T reasoning performance. These results demonstrate that explicitly learning to utilize cross-modal relations is an effective direction for building more accurate and robust RGB-T MLLMs.
\bibliography{aaai2027}

\end{document}